\documentclass{bmvc2k}

\title{Part Segmentation and Motion Estimation for Articulated Objects with Dynamic 3D Gaussians}

\addauthor{Jun-Jee Chao}{chao0107@umn.edu}{1}
\addauthor{Qingyuan Jiang}{jian0345@umn.edu}{1}
\addauthor{Volkan Isler}{isler@cs.utexas.edu}{2}

\addinstitution{
 Department of Computer Science\\
 University of Minnesota\\
 Minnesota, USA
}
\addinstitution{
 Department of Computer Science\\
 The University of Texas at Austin\\
 Texas, USA
}

\runninghead{Chao ET AL.}{Articulated Object Part Segmentation}

\usepackage{times}
\usepackage{comment}
\usepackage{epsfig}
\usepackage{graphicx}
\usepackage{amsmath,amssymb,amsfonts,amsthm,bm}
\usepackage{xcolor}
\usepackage{multicol}
\usepackage{multirow}
\usepackage{booktabs}
\usepackage{array}
\usepackage{algorithm}
\usepackage[noend]{algpseudocode}
\usepackage{makecell}
\usepackage{paralist}
\usepackage{soul}
\usepackage{wrapfig}
\usepackage{epsfig}
\usepackage{graphicx}
\usepackage{adjustbox}
\usepackage{cuted}
\usepackage{capt-of}
\usepackage{tikz} 

\def\eg{\emph{e.g}\bmvaOneDot}

\newcommand{\mymat}[1]{\mathbf{#1}}
\newcommand{\myset}[1]{\mathcal{#1}}
\newcommand{\mypcd}[1]{\textrm{\upshape #1}}

\DeclareMathOperator*{\argmax}{arg\,max}

\newcommand{\secref}[1]{Section~\ref{#1}}
\newcommand{\tabref}[1]{Table~\ref{#1}}

\newcommand{\figref}[1]{Fig.~\ref{#1}}

\begin{document}

\maketitle


\begin{abstract}
Part segmentation and motion estimation are two fundamental problems for articulated object modeling. 
In this paper, we present a method to solve these two problems jointly from a sequence of observed point clouds of a single articulated object.
The main challenge in our problem setting is that the point clouds are not assumed to be generated by a fixed set of moving points. Instead, each point cloud in the sequence could be an arbitrary sampling of the object surface at that particular time step. Such scenarios occur when the object undergoes major occlusions, or if the dataset is collected using measurements from multiple sensors asynchronously. In these scenarios, methods that rely on tracking point correspondences are not appropriate.
We present an alternative approach by representing the object as a collection of simple building blocks modeled as 3D Gaussians. 
With our representation, part segmentation is achieved by assigning the observed points to the Gaussians. Moreover, the transformation of each point across time can be obtained by following the poses of the assigned Gaussian. 
Experiments show that our method outperforms existing methods that solely rely on finding point correspondences.
Additionally, we extend existing datasets to emulate real-world scenarios by considering viewpoint occlusions. We demonstrate that our method is more robust to missing points as compared to existing approaches on these challenging datasets, even when some parts are completely occluded in some time-steps.  Notably, our part segmentation outperforms the state-of-the-art method by $13\%$ on occluded point clouds. Project page: \url{https://giles200619.github.io/gsart_website/}
\end{abstract}

\section{Introduction}
\vspace{-5pt}

\begin{figure}
\includegraphics[width=0.99\linewidth,trim={0.0cm 19cm 0cm 4.3cm},clip]{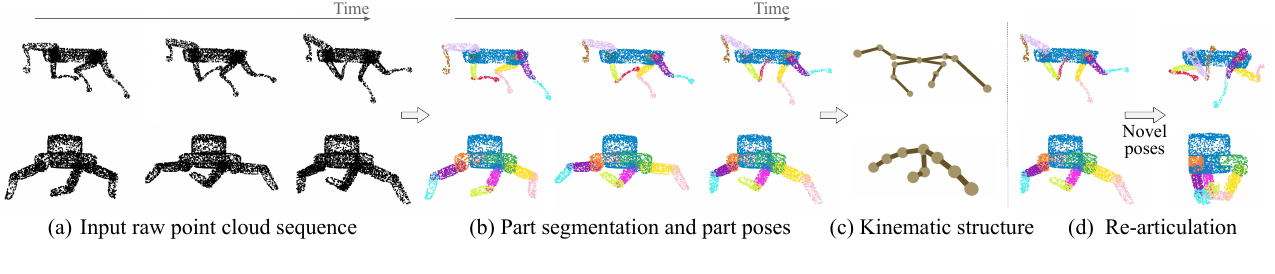}
\vspace{-18pt}
\caption{Problem formulation: (a)~Given a dynamic point cloud sequence of an articulated object, (b-c)~the goal is to jointly estimate per-frame part segmentation, part poses, and the kinematic structure with joint parameters. (d)~The estimated model can both reconstruct the observed motion and re-articulate the object to novel, unseen poses. Note that our method makes no assumptions about object category, kinematic structure, or number of parts.}
\label{fig:teaser}
\vspace{-5pt}
\end{figure}

Identifying rigid parts and motions of articulated objects is fundamental for building digital twins~\cite{jiang2022ditto}, object re-posing~\cite{noguchi2022watch}, and system identification in robotics~\cite{Hsu2023DittoITH,wang2022adaafford}. 
However, building articulated models from point cloud sequences is challenging. 
In real-world scenarios, such sequences are usually captured from multiple sensors asynchronously. This data collection process leads to the main challenge: the observed point cloud sequence is not generated by a fixed set of moving points. Instead, each point cloud in the sequence could be an arbitrary sampling of the object surface at that particular time step. Therefore, corresponding points might not exist across the observed sequence. 
The problem becomes even harder under occlusions caused by varying sensor locations at each time step. 
With only a single sensor, different parts of the object become invisible in different time steps. Hence, methods that rely solely on point correspondences often struggle under these conditions~\cite{huang2021multibodysync,liu2023building}. 
 
Instead of relying solely on point correspondences, we propose an alternative approach to jointly estimate a) part segmentation, b) kinematic tree, and c) joint poses directly from a raw point cloud sequence of a single articulated object (\figref{fig:teaser}). 
Our method represents the articulated object as a set of dynamic 3D Gaussians, where each Gaussian models the motion and point distribution of a rigid part.
A key component of our method is the Gaussian parametrization technique inspired by the success of Gaussian Splatting~\cite{kerbl3Dgaussians,luiten2024dynamic}, where we parameterize each Gaussian with time-varying rotations and translations, while sharing scales across time.
This compact representation enables efficient part segmentation, motion estimation, and point transformation with minimal parameters.
Moreover, the estimated parameters can be used to re-articulate the object to a new pose that has not been seen before.



Unlike many existing methods, we do not make any assumption about the object category~\cite{li2020category,SMPL:2015,chao2023category,liu2023self, zuffi20173d, hani20233d}, number of parts~\cite{jiang2022ditto,liu2023paris, heppert2023carto,tseng2022cla, jain2021screwnet, abbatematteo2019learning}, or the underlying
kinematic tree structure~\cite{wu2022casa, yao2022lassie, NEURIPS2023_a99f50fb, wu2023magicpony}, therefore making our method more generalizable to in-the-wild objects and more suitable for real-world applications. 
We show in \secref{sec:exp} that our method outperforms existing works on two established benchmarks. Furthermore, while existing approaches focus on point clouds that fully cover the object surface, less attention has been given to modeling point cloud sequences with occlusions. To assess performance under such conditions, we extend the dataset to mimic real-world scenario by involving viewpoint occlusions. Experiment results show that our method is more robust to missing points in the observed point clouds compared to existing works. Our contributions can be summarized as:
\begin{compactitem} 
\item  To build an articulated model from raw 4D point cloud of an arbitrary articulated object, we present a compact and effective representation by modeling rigid part segmentation and motion with dynamic 3D Gaussians. 
\item Compared to most existing point cloud-based methods that solely rely on point correspondence, our method that considers point distribution outperforms existing methods on established benchmarks. 
\item 
We study the performance of our method when the input point clouds undergo occlusions. Experiments show that our method is more robust to missing points than existing approaches. Notably, our part segmentation outperforms the state-of-the-art (SoTA) method by $13\%$ on point clouds with partial observations.
\end{compactitem}


\section{Related Work}

\textbf{Object part segmentation.}
Many methods perform part segmentation on static point clouds using deep learning~\cite{qi2017pointnet++,qian2022pointnext,thomas2019kpconv,wang2019dynamic,yi2019gspn,liu2019relation,liu2023partslip,liu2023self, liu2023semi}. Others extend part segmentation to dynamic sequences by leveraging motion cues~\cite{hayden2020nonparametric,yi2018deep,deng2024banana}. 
More recently, a line of research has evolved to jointly perform part segmentation and motion estimation from point cloud observations~\cite{shi2021self,abdul2022learning, huang2021multibodysync, nie2023structure}.
MultiBodySync~\cite{huang2021multibodysync} introduces a differentiable permutation and segmentation module to iteratively find the optimal matching between point clouds. 
However, these methods often assume all parts are always completely visible. In contrast, we demonstrate that our method works on both complete and partial point cloud sequences.


\textbf{Articulated object modeling.}
A major line of research reconstructs articulated shapes from images or videos with the help of informative visual features~\cite{wu2022casa,yang2022banmo,song2024reacto,yang2021lasr, wei2022nasam, lei2024gart}. However, some of these works focus on single object categories like humans~\cite{SMPL:2015,MANO:SIGGRAPHASIA:2017} or animals~\cite{zuffi20173d,wu2022casa, yao2022lassie, NEURIPS2023_a99f50fb, wu2023magicpony}, where a kinematic structure is shared between all instances. 
For instance, GART~\cite{lei2024gart} uses a categorical template model to recover the targets' appearance from monocular videos. 
With strong categorical prior, these methods often focus on surface reconstruction and novel view synthesis, rather than part segmentation and motion analysis.

Some category-agnostic methods make assumptions about the number of moving parts~\cite{jiang2022ditto,liu2023paris, heppert2023carto,tseng2022cla, jain2021screwnet, Kuai_2023_CVPR}. For example, Ditto~\cite{jiang2022ditto} and PARIS~\cite{liu2023paris} focus on objects with a single moving part given two observed articulated states. 
More recent works aim to generalize to arbitrary articulated objects.
Watch-It-Move~\cite{noguchi2022watch} models part poses and surfaces with ellipsoids and signed-distance functions (SDFs), which are optimized via volumetric rendering~\cite{wang2021neus}. Then the underlying kinematic structure is inferred from the estimated part poses.

The closest work is Reart~\cite{liu2023building}, which proposes a two-step optimization approach: a relaxed model with independent part poses and segmentation is first fitted to the observed 4D point cloud without any kinematic constraints. 
Then the underlying kinematic structure is estimated from the relaxed model. 
However, Reart selects a canonical time step and fits a MLP to segment points at this time step. Therefore, their segmentation model does not generalize to point clouds at other time steps, and the performance is sensitive to the chosen canonical frame. In contrast, we model the surface point distribution and part motion with 3D Gaussians, allowing us to segment point clouds at all time steps. 

\vspace{-3pt}
\section{Method}
\label{sec:method}

 Our goal is to estimate rigid part segmentation, kinematic structure and joint states from a sequence of $K$ 3D point clouds $\myset{X}=\{\mypcd{X}^k\}_{k \in 1,...,K}$, where each point cloud $\mypcd{X}^k=[x^k_1,...,x^k_n,...,x^k_N]$ consists of $N$ points in $\mathbb{R}^{3}$ sampled from the surface of an articulated object. Each point $x$ belongs to one of the rigid parts of the object. Note that each point cloud $\mypcd{X}^k$ is arbitrarily sampled at every time step $k$, therefore point correspondences do not exist in $\myset{X}$, since the corresponding points of $x^k_i$ might not be sampled in the other time steps. 
 
 To estimate an articulated model from the given set of point cloud  $\myset{X}$, we propose an optimization approach that consists of 3 steps.
 First, we optimize $m$ 3D Gaussians independently where each Gaussian represents the point distribution and motion of a single rigid part (\secref{sec:method_gaussian}). 
Then we build a kinematic tree by adding edges between part pairs whose relative poses are close to 1-DOF motion (\secref{sec:method_tree}).
 Finally, the joint parameters of the kinematic tree are fine-tuned in a kinematic chain-aware fashion (\secref{sec:kinematic_fine_tune}). 
 \begin{figure}[t!]
    \centering
    \includegraphics[width=\linewidth,trim={0cm 17cm 0cm 3.5cm},clip]{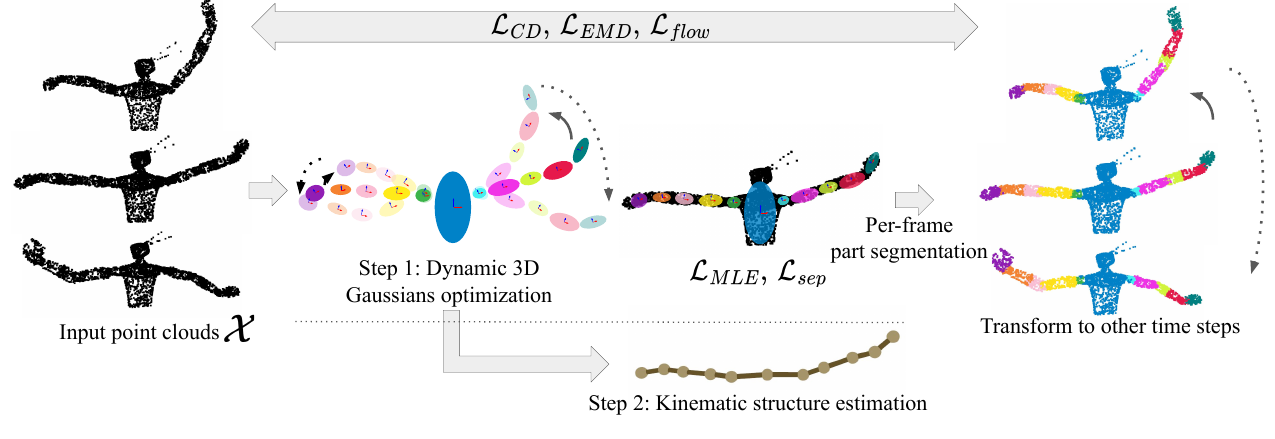}
    \vspace{-25pt}
    \caption{Our optimization pipeline. In each iteration, we sample a point cloud from $\myset{X}$ uniformly at random, and optimize the Gaussian parameters at that time step by minimizing $\mathcal{L}_{MLE}$ and $\mathcal{L}_{sep}$ to maximize the log-likelihood of the observed point cloud, and ensure each Gaussian represents a single rigid part. Then we assign points to the Gaussians with the smallest Mahalanobis distance. Finally, we transform the segmented point cloud to other time steps with the corresponding Gaussian poses, and enforce similarity with the observed point clouds at those time steps.  }
    \label{fig:method}
    \vspace{-0pt}
\end{figure}

\vspace{-0pt}
\subsection{Step 1: Modeling rigid parts with dynamic 3D Gaussians}
\label{sec:method_gaussian}

\paragraph{Gaussian parametrization}

 Each Gaussian $G$ is defined by a full 3D covariance matrix $\Sigma$ and center $\mu$ in the world frame: $G(x) = \frac{1}{(2\pi)^{n/2}|\Sigma|^{1/2}} exp(-\frac{1}{2}(x-\mu)^T\Sigma^{-1}(x-\mu))$. The covariance matrix $\Sigma$ is decomposed into a diagonal scaling matrix $\mymat{S}$ and rotation $\mymat{R}$ : $\Sigma = \mymat{R}\mymat{S}(\mymat{R}\mymat{S})^{T}$. 
This representation allows independent optimization of the three factors for each Gaussian: a center $\mu \in \mathbb{R}^{3}$, scale $s \in \mathbb{R}^{3}$, and rotation matrix $\mymat{R} \in SO(3)$ represented with the 6D representation~\cite{zhou2019continuity}. Moreover, the transformation $\mymat{T} \in SE(3)$ that transforms points from the Gaussian local frame to the observed point cloud frame (the world frame) can be defined as: $\mymat{T} = \begin{bmatrix}
   \mymat{R}   & \mu\\
    0     & 1
\end{bmatrix}$.
For each $G$ to model the part motion over $K$ time steps, we optimize $K$ different $\mymat{R}$ and $\mu$ while sharing the same $s$ across all time steps. 
We denote the $i$-th Gaussian at the $k$-th time step as $G^k_i$, defined by $\mymat{R}^k_i$, $\mu^k_i$, and $s_i$, with the associated part pose $\mymat{T}^k_i$.

\paragraph{Rigid part segmentation and point transformation}
We design the Gaussians to simultaneously encode both motions and point distributions of the rigid parts, allowing us to perform part segmentation without extra parameters. 
Given a 3D point $x^k$ at time $k$, part segmentation is naturally achieved by selecting the Gaussian with the smallest Mahalanobis distance:
\begin{equation}
\label{eq:segmentation}
f(x^k) = \argmax_{i \in \{1, \dots, m\}}(-\mathcal{M}(x^k, G^k_i))
\end{equation}
where $\mathcal{M}$ is the squared Mahalanobis distance:${(x^k-\mu^k_i)^T{\Sigma^k_i}^{-1}(x^k-\mu^k_i)}$.
To maintain differentiability during optimization, we replace $\argmax$ with the Gumbel-Softmax trick~\cite{jang2017categorical,maddison2017concrete}.

With our representation, point cloud $\mypcd{X}^k$ can be transformed from time $k$ to time $t$:

\begin{equation}
\mathcal{H}_{k \rightarrow t} = \hat{\mypcd{X}^t} = \{\mymat{T}^t_{f(x)} \cdot {\mymat{T}^k_{f(x)}}^{-1} \cdot \begin{bmatrix} x \\ 1 \end{bmatrix} \mid x\in \mypcd{X}^k\}
\end{equation}
The points are first transformed to the local frames of the estimated parts, then transformed back to the world frame with the part poses at other time steps. Moreover, we can fuse the observed point clouds from all time steps to a single time step $t$ after transforming to the same object pose: $\bigcup_{k=1}^K \mathcal{H}_{k \rightarrow t}$. We find this fusion ability critical, especially when handling point clouds with occlusions, since it allows us to gather information from all time steps. 

\paragraph{Gaussian optimization}
The Gaussian parameters are optimized in an analysis-by-synthesis fashion as illustrated in \figref{fig:method}. In every iteration, we sample a point cloud $\mypcd{X}^k$ from $\myset{X}$ uniformly at random and perform a gradient step to minimize the loss: 
$\mathcal{L} = \lambda_{MLE}\mathcal{L}_{MLE} + \lambda_{sep}\mathcal{L}_{sep} + \lambda_{CD}\mathcal{L}_{CD} + \lambda_{EMD}\mathcal{L}_{EMD} + \lambda_{flow}\mathcal{L}_{flow}$.
The goal is to enforce all $m$ Gaussians to jointly model the observed point distributions while individual Gaussian represents a single rigid part. 
Moreover, $\mypcd{X}^k$ at time step $k$ should be similar to any other point cloud $\mypcd{X}^t$ after being transformed with the corresponding Gaussian poses. 
Next, we detail each loss term:

\textbf{Maximum likelihood loss:} Given $\mypcd{X}^k$, $\mathcal{L}_{MLE}$ optimizes the Gaussian parameters to maximize the log-likelihood of the observed points. This loss is similar to the M step in the EM algorithm~\cite{moon1996expectation} for fitting a Gaussian Mixture Model (GMM)~\cite{reynolds2009gaussian}. However, unlike standard GMMs, we enforce equal weights for all Gaussians to ensure that each Gaussian represents a single rigid part, while the overall Gaussians cover the entire point cloud.

\begin{equation} 
\mathcal{L}_{MLE}  = -\sum_{n=1}^N log(\sum_{i=1}^m G_i^k(x_n^k) )
\end{equation}

\textbf{Separation loss:} To ensure that each Gaussian represents a single rigid part, we apply $\mathcal{L}_{sep}$ to avoid any two Gaussians being too close or overlap with each other.

\begin{equation} 
\mathcal{L}_{sep}  = \frac{1}{m} \sum_{i=1}^m exp(-\alpha \cdot \min_{j\neq i}(\mathcal{M}(\mu_j^k, G_i^k)))
\end{equation}
where $\alpha$ controls the minimum Mahalanobis distance between the center of a Gaussian to another Gaussian distribution. Larger $\alpha$ allows the Gaussians to be closer with each other.

\textbf{Chamfer distance:} Given one observed point cloud $\mypcd{X}^k$, we can transform the points to all time steps ${\{\mathcal{H}_{k \rightarrow t} \,|\, t = [1,...,K]\}}$. We apply $\mathcal{L}_{CD}$ to enforce the transformed point clouds to match the observations at other time steps by minimizing the Chamfer distance.

\begin{equation}
\begin{split}
 \mathcal{L}_{CD} = \sum_{t=1}^K\left(\sum_{x \in \mathcal{H}_{k \rightarrow t}} \min_{y \in \mypcd{X}^t} \|x - y\|_2^2 +  \sum_{y \in \mypcd{X}^t} \min_{x \in \mathcal{H}_{k \rightarrow t}} \|x - y\|_2^2\right)
\end{split}
\end{equation}

\textbf{Earth-mover distance:} Similar to Chamfer distance, $\mathcal{L}_{EMD}$ measures the similarity between two point clouds. Instead of finding the closest point as correspondence, $\mathcal{L}_{EMD}$ finds the optimal matching between two point clouds by solving a bipartite matching problem~\cite{jonker1987shortest}.

\begin{equation}
\mathcal{L}_{EMD} = \sum_{t=1}^K \min_{\mypcd{A}} \|A\mathcal{H}_{k \rightarrow t} - \mypcd{X}^t\|_2^2 
\end{equation}
where $\mypcd{A}$ denotes the permutation matrix solved using the linear assignment solver~\cite{crouse2016implementing}.

\textbf{Flow loss:} As demonstrated in~\cite{huang2021multibodysync,liu2023building}, flow prediction provides important cue regarding motion transition between point clouds. $\mathcal{L}_{flow}$ encourages the point-wise 3D motion generated by our articulated model matches the prediction from a scene flow network.

\begin{equation}
\mathcal{L}_{flow} = \sum_{t=1}^{K-1}\bigg\| (\hat{\mypcd{X}^{t+1}} - \hat{\mypcd{X}^t}) - g(\mypcd{X}^{t+1}, \mypcd{X}^t;\hat{\mypcd{X}^t})\bigg\|_2^2 
\end{equation}
where $g$ is a pre-trained scene-flow network from~\cite{huang2021multibodysync,liu2023building} that estimates the point-wise motion $g(\mypcd{X}^{t+1}, \mypcd{X}^t;\hat{\mypcd{X}^t})$ for all points in $\hat{\mypcd{X}^t}$ given two observed point clouds.

Another important factor is selecting an appropriate number of parts $m$ during initialization. 
In most clustering algorithms, $m$ is a hyperparameter chosen empirically using model selection criteria such as AIC for EM~\cite{steele2010performance}.
We follow a similar approach: we run our optimization with various values of $m$ and choose the one that yields the minimum $\mathcal{L}_{CD}$ after convergence. Additional detail about how this parameter affects our performance can be found in the supplementary material.

\subsection{Step 2: Kinematic tree estimation}
\label{sec:method_tree}

Given the predicted parts and motions, we adapt a scheme similar to~\cite{liu2023building,noguchi2022watch} for estimating the kinematic tree where each edge connects two parts with 1-DOF relative motion. We briefly outline the process here and provide more detail in the supplementary material. 

If two parts form a parent-child pair in the kinematic tree, they must be spatially close, and  their relative poses at all time steps should share the same rotation or translation axis. These two properties are measured with $\mathcal{L}_{spatial}$ and $\mathcal{L}_{1-DOF}$, respectively. 
For part $i$ and $j$, $\mathcal{L}_{spatial}$ measures the shortest Euclidean distance between any pair of points: $\mathcal{L}_{spatial} = \min_{x \in \mypcd{X}_i} \min_{y \in \mypcd{X}_j} \big\|x - y \big\|_2^2$,
where $\mypcd{X}_i$ and $\mypcd{X}_j$ denote the points that are segmented as the $i$-th and $j$-th part respectively. 
$\mathcal{L}_{1-DOF}$ measures how well the relative motion between two parts can be approximated by a 1-DOF relationship: $\mathcal{L}_{1-DOF} = \sum_k \bigg\| ({\mymat{O}_i^k}^{-1}  \cdot \mymat{O}_j^k) \cdot {\mymat{S}_{ij}^k}^{-1} - \mymat{I} \bigg\|_{F}^2$,
where $\mymat{I}$ denotes the identity matrix. Here, $\mymat{O}^k_i = {\mymat{T}_i^{k+1}} \cdot {\mymat{T}_i^{k}}^{-1} $ is the motion of part $i$ in the world frame between consecutive time steps. Hence, $({\mymat{O}_i^k}^{-1}  \cdot \mymat{O}_j^k)$ calculates the relative motion of the two parts.  ${\mymat{S}_{ij}^k}$ is the approximated 1-DOF relative motion between the two parts, considering their relative poses across all time steps. Essentially, $\mathcal{L}_{1-DOF}$ measures the deviation of the observed relative motion from the approximated ideal 1-DOF motion.

Before constructing the kinematic tree, we first merge parts that are spatially close and their relative motion is close to static across all time steps. 
Then we compute pairwise $\mathcal{L}_{spatial}$ and $\mathcal{L}_{1-DOF}$ between the remaining parts after merging. Finally, the kinematic tree is built by solving the minimum spanning tree problem (MST)~\cite{cormen2022introduction} such that the total loss: $\lambda_{spatial}\mathcal{L}_{spatial} + \lambda_{1-DOF}\mathcal{L}_{1-DOF}$ of all edges in the tree is minimized.

\subsection{Step 3: Joint parameters fine-tuning with kinematic constraints}
\label{sec:kinematic_fine_tune}

Finally, the kinematic chain is constructed by first setting the part with the least motion across all time steps as the root node, followed by traversing outward to all leaf nodes to build the parent-child relationships.
Joint parameters are then estimated by projecting Gaussian poses onto the chain, approximating 1-DOF screw parameters~\cite{lynch2017modern,stramigioli2001geometry} for each parent-child pair. These parameters are further refined via forward kinematics by minimizing $\mathcal{L}_{CD}$ and $\mathcal{L}_{EMD}$. 

\section{Experiments}
\label{sec:exp}


\begin{figure}
\begin{tabular}{cc}
\bmvaHangBox{
\includegraphics[width=0.45\linewidth,trim={0.3cm 9cm 3.5cm 3.2cm},clip]{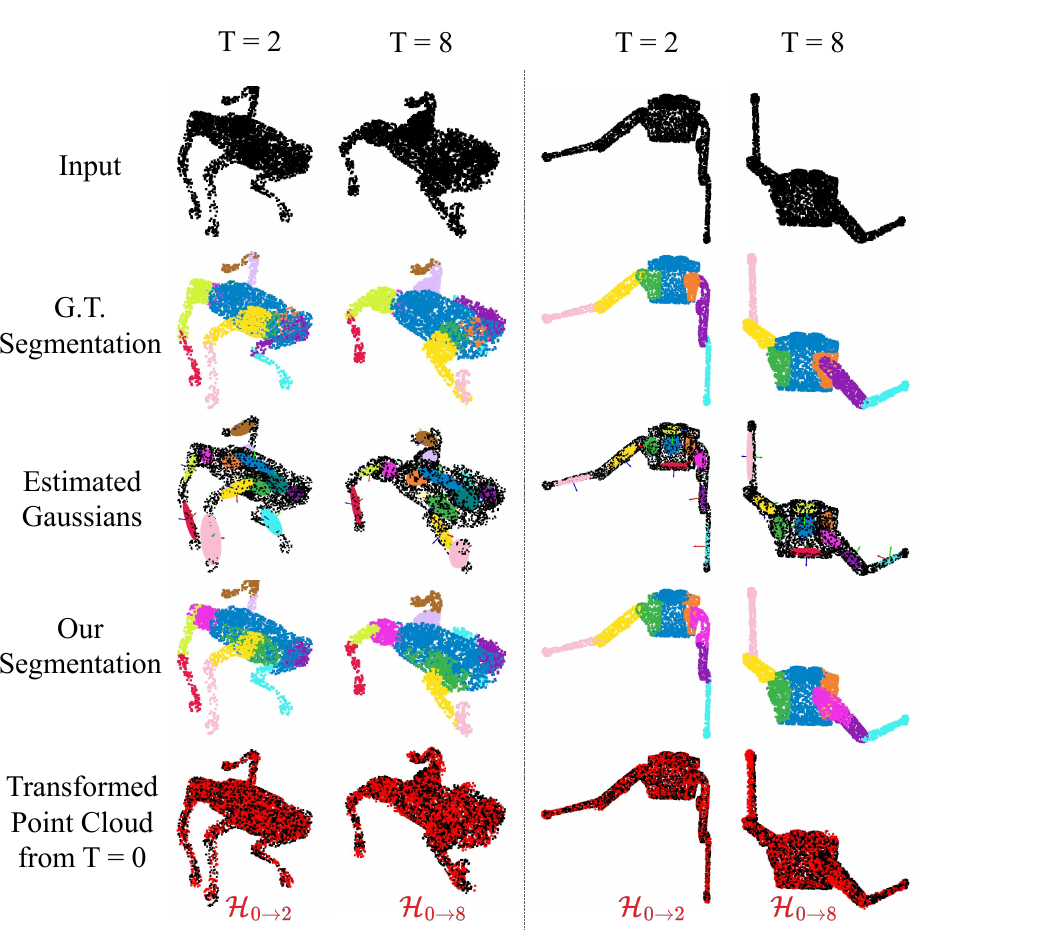}
} &\hspace{-38pt} {\vrule height 0.1cm depth 5.2cm}
\bmvaHangBox{
\includegraphics[width=0.6\linewidth,trim={0.7cm 10.5cm 2.3cm 3.3cm},clip]{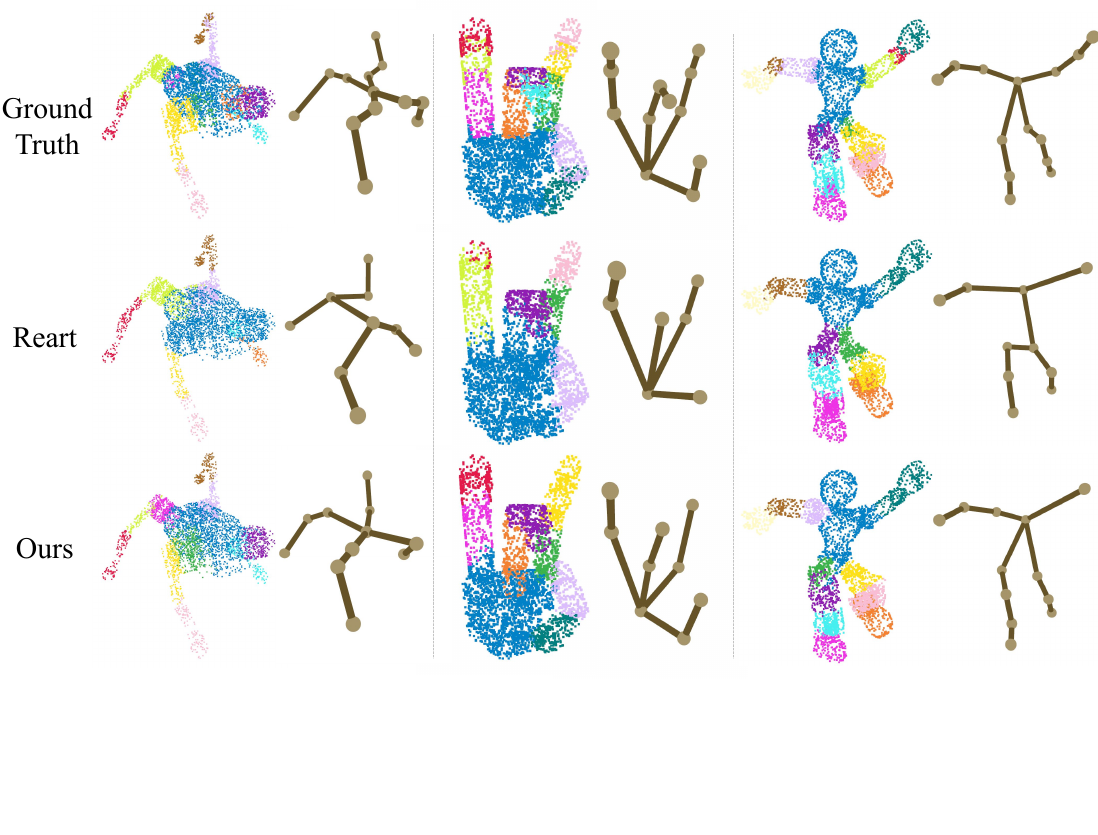}
} \\[-20pt]
(i)&(ii)
\end{tabular}
\vspace{-10pt}
\caption{Qualitative results on the RoboArt dataset. (i) The third row shows our estimated Gaussians overlaid with the input point cloud (black). Segmentation results in the fourth row are obtained by assigning the points to the closest Gaussians, followed by the merging process in~\secref{sec:method_tree}. The last row demonstrates motion quality by overlapping the observed point cloud at the current time (black) with points transformed from $\text{T}=0$ (red), showing that the motion is well reconstructed even though they are two different point sets. (ii) Compared to Reart~\cite{liu2023building}, our method is able to capture smaller parts more faithfully, for instance, the knuckles of the robot hand, leading to more accurate kinematic tree estimation.}
\label{fig:quali_roboart}
\vspace{-0pt}
\end{figure}

We evaluate our method on part segmentation, motion estimation, and re-articulation accuracy in \secref{sec:exp_roboart}. We then assess segmentation performance on a larger dataset of daily objects in \secref{sec:exp_sapien}. Additionally, we create challenging datasets with occlusions by considering point visibility and demonstrate our method's capability of handling noisy data in \secref{sec:exp_partial}, even when some parts are completely occluded in some time-steps.

\subsection{Experimental Setup}
\label{sec:exp_setup}

\textbf{Datasets:} 
Experiments are conducted on two main datasets: the \textbf{RoboArt} dataset~\cite{liu2023building}, containing robot instances with up to 15 articulated parts and diverse kinematic structures (e.g., robot fingers, quadrupeds), and the \textbf{Sapien} dataset~\cite{xiang2020sapien,huang2021multibodysync}, which includes 720 test sequences across 20 categories of daily objects. Note that the point clouds are arbitrarily resampled at every time step. 
To test robustness to occlusions, we generate two RoboArt variants: \textbf{Partial-RoboArt}, where only self-occlusion within each rigid part is considered (ensuring all parts remain partially visible), and \textbf{Occluded-RoboArt}, where we simulate real occlusions by removing points not visible from an arbitrarily sampled camera at each time step~\cite{katz2007direct,Zhou2018}, making some parts completely invisible in certain frames.

\textbf{Metrics:} 
\textbf{Reconstruction Error} and \textbf{Flow Error}~\cite{liu2023building,huang2021multibodysync} measures the mean squared error (MSE) and point flow difference between point cloud sequences transformed with the ground truth vs.\ estimated parameters. 
\textbf{Rand Index (RI)}~\cite{chen2009benchmark, liu2023building} measures segmentation accuracy by calculating the pairwise agreement between the ground truth and predicted part label. \textbf{Tree Edit Distance}~\cite{pawlik2015efficient,zhang1989simple} measures how close the predicted kinematic tree is to the ground truth tree. Finally, \textbf{Reanimate Error}~\cite{liu2023building} measures the MSE between the unseen articulated point clouds and the predicted reanimated point clouds to unseen poses. 



\subsection{Result on modeling articulated objects}
\label{sec:exp_roboart}

\begin{table}[t]
  \centering
  \begin{minipage}[b]{0.5\textwidth}
    \centering
    \resizebox{1\columnwidth}{!}{\begin{tabular}{cccccc}
\toprule
Method        & \begin{tabular}[c]{@{}c@{}}Recons\\ Error$\downarrow$\end{tabular} & \begin{tabular}[c]{@{}c@{}}Rand\\ Index$\uparrow$\end{tabular} & \begin{tabular}[c]{@{}c@{}}Tree Edit\\ Distance$\downarrow$\end{tabular} & \begin{tabular}[c]{@{}c@{}}Reanimate\\ Error$\downarrow$\end{tabular} & \begin{tabular}[c]{@{}c@{}}Flow\\ Error$\downarrow$\end{tabular} \\ 
\midrule
MultiBodySync~\cite{huang2021multibodysync} & 4.76                                                   & 0.70                                                 & 6.6                                                          & 9.72                                                      & 3.42                                                 \\
WatchItMove~\cite{noguchi2022watch} & 12.77                                                  & 0.78                                                 & 6.6                                                          & 7.43                                                      & 8.42                                                 \\
Reart~\cite{liu2023building} & 1.26                                                   & 0.86${}^\ast$                                                & 2.9                                                          & 3.66                                                      & 0.57                                                 \\
Ours${}^\dagger$ & 1.17 & \textbf{0.90} & \textbf{2.7} & - & 0.96 \\
Ours          & \textbf{0.88} & \textbf{0.90}   & \textbf{2.7}   & \textbf{2.93}   & \textbf{0.48}                                                
\end{tabular}}
    \vspace{-8pt}
    \caption{Results on the RoboArt test set. ${}^\dagger$ denotes running our method without kinematic parameter fine-tunning (\secref{sec:kinematic_fine_tune}). ${}^\ast$ Note that Reart evaluates their segmentation performance only on a single canonical frame, while we evaluate our method on all frames.}
    \label{table:reart_test}
  \end{minipage}
  \hfill
  \begin{minipage}[b]{0.48\textwidth}
    \centering
    \resizebox{1.0\columnwidth}{!}{\begin{tabular}{cccc}
\toprule
Method        & Flow Error$\downarrow$ & Multi-scan RI$\uparrow$ & Per-scan RI$\uparrow$ \\ 
\midrule
PWC-Net~\cite{wu2020pointpwc} & 6.20       & -             & -           \\
PointNet++~\cite{qi2017pointnet++} & -          & 0.62          & 0.65        \\
MeteorNet~\cite{liu2019meteornet} & -          & 0.59          & 0.60        \\
Deep Part~\cite{yi2018deep} & 5.95       & 0.64          & 0.67        \\
NPP~\cite{hayden2020nonparametric} & 21.22      & 0.63          & 0.66        \\
MultiBodySync~\cite{huang2021multibodysync} & 5.03       & 0.76          & 0.77        \\
Reart~\cite{liu2023building} & 4.79 & 0.79  & 0.79 \\
Ours          & 4.63 & 0.79  & 0.81 \\
Ours${}^\ast$          & \textbf{4.58} & \textbf{0.82}  & \textbf{0.83}      
\end{tabular}}
    \vspace{-5pt}
    \caption{Results on the Sapien dataset. $^\ast$denotes selecting the best result among the three initial parameters ($m = 3,5,7$).}
    \label{table:sapien}
  \end{minipage}
\end{table}

We visualize the estimated 3D Gaussians in \figref{fig:quali_roboart}-(i), showing that each Gaussian follows the motion of its rigid part across time. We evaluate the estimated motion qualitatively in the last row. Despite the input point cloud sequence is sampled arbitrarily at each time step, the union of the observed and the transformed point clouds cover the object surface faithfully.
As shown in \tabref{table:reart_test}, our method outperforms existing approaches in all metrics on the RoboArt test set. 
Unlike correspondence-based methods~\cite{huang2021multibodysync, liu2023building}, which may fail when correspondences do not exist, our method models each rigid part as a 3D Gaussian and enforces similarity in point distributions across time for points belonging to the same part. As demonstrated in \figref{fig:quali_roboart}-(ii), this enables finer motion capture and more accurate segmentation of small parts, such as the knuckles of the robot hand or humanoid's shoulder, which the baseline fails to achieve. 
In this experiment, we run our method with $m$ from $10$ to $15$ for each instance and select the one with the least $\mathcal{L}_{CD}$ after optimization. Similar process is employed in existing methods, for example, Reart~\cite{liu2023building} determines the canonical frame by performing multiple optimizations with different time step as the canonical frame and selecting the one with the lowest loss after convergence.
Notably, Reart segments points only in the canonical time step, while our method is capable of segmenting points across all frames. Even when segmentation performance is evaluated over all time steps, as opposed to Reart's single-frame segmentation, our method achieves better result, as presented in \tabref{table:reart_test}.

\subsection{Result on segmenting object parts}


\label{sec:exp_sapien}
We further evaluate the performance on part segmentation and flow prediction with the Sapien dataset (\figref{fig:sapien}), which contains man-made objects with fewer articulated parts than RoboArt. Therefore we run our method with $m = 3,5,7$ as the number of part initialization. 
In addition to Per-scan RI, which evaluates segmentation frame-by-frame, we also include Multi-scan RI that assesses segmentation consistency across time.
As shown in \tabref{table:sapien}, optimization-based methods (\eg, Reart~\cite{liu2023building}, MultiBodySync~\cite{huang2021multibodysync}, Deep Part~\cite{yi2018deep}) outperform feed-forward methods (\eg, PointNet++~\cite{qi2017pointnet++}, PWC-Net~\cite{wu2020pointpwc}) by trading off inference speed for better accuracy. 
Moreover, our per-frame segmentation performance outperforms MultiBodySync~\cite{huang2021multibodysync} that seeks to find point correspondences and Reart~\cite{liu2023building} that solves part-segmentation in a single frame. We also present the upper bound of our performance in \tabref{table:sapien}, by reporting the best result among the three initial $m$. We show that with a better parameter selecting criterion, the performance of our method can be further improved.

\subsection{Result on partial point clouds}
\label{sec:exp_partial}

We compare our method with Reart~\cite{liu2023building}, which models part segmentation with a coordinate-based MLP, on two challenging datasets: Partial-RoboArt and Occluded-RoboArt. These datasets simulate real-world scenarios with occlusions, where not all parts are fully visible across time. Both methods are initialized with the same number of parts ($m=20$). To handle missing points, we modify $\mathcal{L}_{CD}$ and $\mathcal{L}_{EMD}$ to compute the one-directional matching from the observed point cloud $\mypcd{X}^t$ to the fused point cloud from other time steps: $\bigcup_{k=1, k\neq t}^K \mathcal{H}_{k \rightarrow t}$. Details are provided in the supplementary material.

\begin{figure}[t]
  \centering
  \begin{minipage}[b]{0.47\textwidth}
    \centering
    \includegraphics[width=0.99\linewidth,trim={2.7cm 11cm 3cm 3cm},clip]{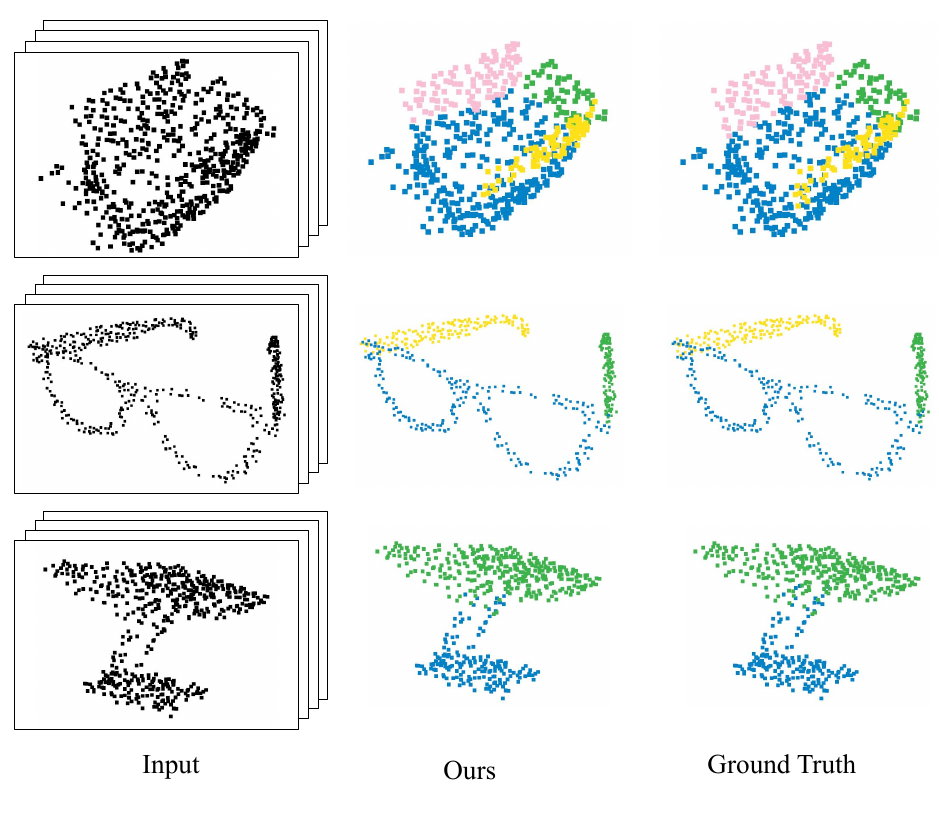}
    \vspace{-10pt}
    \caption{Qualitative result of our part segmentation on the Sapien dataset.}
    \label{fig:sapien}
  \end{minipage}
  \hfill
  \begin{minipage}[b]{0.52\textwidth}
    \centering
    \includegraphics[width=\linewidth,trim={0.1cm 14.5cm 10.8cm 3cm},clip]{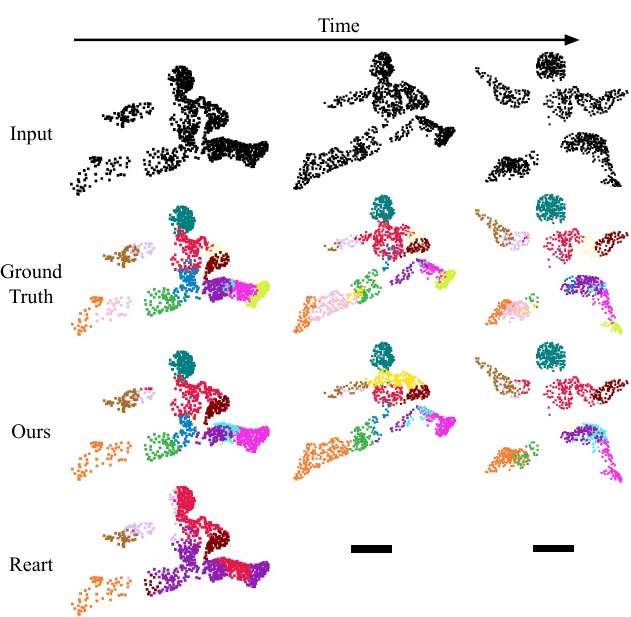}
    \vspace{-20pt}
    \caption{Qualitative results on Occluded-RoboArt. Our method separates the parts where the baseline fails.} 
    \label{fig:occlusion}
  \end{minipage}
\end{figure}

In Partial-RoboArt, all parts are partially observed across time, while some parts are fully missing in some frames in Occluded-RoboArt. As shown in \tabref{table:partial}, both methods perform worse when there are more missing points, while our method is less affected.
\figref{fig:occlusion} shows that our method separates parts more reliably than Reart. However, due to missing information, our method can produce noisy or inconsistent segmentation (\eg, the yellow part in the middle frame of the humanoid). This can happen when our method  accidentally fits multiple Gaussians to different portions of that rigid part.
Notably, Reart's reliance on single-frame segmentation makes it more vulnerable when points do not exist in its selected canonical frame but appear in other time steps. 
This experiment highlights the advantages of considering point distribution of the rigid parts, and demonstrates that our method is more robust to occluded point clouds or points that do not completely cover the object surface.

\subsection{Ablation studies}
\label{sec:ablation}

\begin{table}[t]
  \centering
  \begin{minipage}[b]{0.52\textwidth}
    \centering
    \resizebox{1\columnwidth}{!}{
    \begin{tabular}{lcc|cc}
        \toprule
        \multirow{2}{*}{} & \multicolumn{2}{c|}{Partial-RoboArt} & \multicolumn{2}{c}{Occluded-RoboArt} \\ 
        \cmidrule(lr){2-3} \cmidrule(lr){4-5}
                               & Reart~\cite{liu2023building} & Ours & Reart~\cite{liu2023building} & Ours \\ 
        \midrule
        Recons Error $\downarrow$       & 4.65     & \textbf{1.99}    & 7.66     & \textbf{2.68}     \\
        Cano Frame RI $\uparrow$        & 0.78     & \textbf{0.85}     & 0.75     & \textbf{0.85}     \\
        Mean RI $\uparrow$           & -     & \textbf{0.85}     & -     & \textbf{0.83}     \\
        Flow Error $\downarrow$   & 2.64  & \textbf{1.26}    & 4.73     & \textbf{1.74}   \\
        Tree Edit Distance $\downarrow$   & 5.5    & \textbf{4.8}   & 6.3     & \textbf{5.4}  \\
        \bottomrule
        \end{tabular}
    }
    \captionof{table}{Results on Partial-RoboArt and Occluded-RoboArt test set. Both datasets consist of point clouds with missing points, while Occluded-RoboArt has more missing parts. Result shows that Reart suffers more when there are more missing points, while our method is less sensitive to missing information.}
    \label{table:partial}
  \end{minipage}
  \hfill
  \begin{minipage}[b]{0.46\textwidth}
    \centering
    \resizebox{1\columnwidth}{!}{\begin{tabular}{ccccc}
\toprule
Loss terms       & \begin{tabular}[c]{@{}c@{}}Recons\\ Error$\downarrow$\end{tabular} & \begin{tabular}[c]{@{}c@{}}Rand\\ Index$\uparrow$\end{tabular} & \begin{tabular}[c]{@{}c@{}}Flow\\ Error$\downarrow$\end{tabular}  & \begin{tabular}[c]{@{}c@{}}Tree Edit\\ Distance$\downarrow$\end{tabular}\\ 
\midrule
w/o $\mathcal{L}_{MLE}$  & 3.03 & 0.81 & 1.90 & 6.25   \\ 
w/o $\mathcal{L}_{sep}$  &  1.45 & 0.88 & 1.21 & 4.0 \\
w/o $\mathcal{L}_{CD}$   &  1.51 & 0.87 & 1.14 & 4.0 \\
w/o $\mathcal{L}_{EMD}$  &  10.04 & 0.75 & 6.70 & 6.5  \\
w/o $\mathcal{L}_{flow}$  &  14.52 & 0.86 & 16.19 & 4.5 \\
Ours          & \textbf{1.26} & \textbf{0.89}   & \textbf{1.05}   & \textbf{2.75}                                  
\end{tabular}}
\caption{Ablation study on RoboArt validation set. We evaluate the importance of each loss term by removing them one at a time.}
    \label{table:ablation}
  \end{minipage}
\end{table}

We conduct an ablation study to evaluate the effect of each loss term. 
As shown in \tabref{table:ablation}, all components are essential. Among the point cloud reconstruction losses, $\mathcal{L}_{EMD}$ and $\mathcal{L}_{flow}$, are particularly important, since they consider global point matching and motion accuracy. In contrast, $\mathcal{L}_{CD}$ has less impact, which is due to its nearest-neighbor matching can easily result in incorrect correspondences. Moreover, $\mathcal{L}_{MLE}$ significantly affects the performance of part segmentation, since it regularizes the Gaussians to cover the observed point clouds by enforcing equal weights. Without this loss, we observe that the model tends to ignore smaller rigid parts by using a larger Gaussian to approximate the motion of several neighboring smaller parts.




\section{Conclusion}
\label{sec:conclusion}

We presented a novel optimization approach to model arbitrary articulated objects from raw point cloud sequences. Our method jointly estimates per-frame part segmentation, part poses, and the kinematic structure with joint parameters, enabling re-articulation to unseen poses.
By modeling rigid parts as point distributions, 
our method outperforms existing methods that solely rely on point correspondences on established benchmarks. 
Moreover, we created datasets that emulate real-world scenarios by considering viewpoint occlusions. Experiments showed that our method is more robust to missing points compared to past works, even when some parts are completely occluded in some frames. 
Notably, our segmentation achieves up to a $5\%$ improvement over the SoTA on existing datasets, and a $13\%$ improvement under partial observations.
However, our method may fail when a large portion of the object is not seen across multiple frames. One possible solution is to apply a motion-aware generative model to fill in the missing information, which we leave for future work. 

\paragraph{Limitations.}
As presented, our method requires the number of clusters $m$ to be given as input. Although we propose a decision criterion for choosing the best $m$ from multiple initializations, this requires additional computation time.  A better solution, which we plan to explore in the future, could be to train a network for estimating the number of clusters. 
Additionally, our method assumes that the kinematic structure is acyclic and the parent-child part pairs exhibit 1-DOF relative motion. While these assumptions hold for for many real-world objects, the kinematic tree estimation module needs to be generalized when these assumptions fail.

\paragraph{Acknowledgment.}
This work was funded in part by the National Research Foundation of Korea (NRF) grant (MSIT) No. RS-2024-00462874.

\bibliography{references}
\end{document}


\maketitle


In this supplementary material, we provide additional implementation details in \secref{sec:sup_implementation}. In \secref{sec:sup_result}, we first investigate how the number of initialized Gaussians $m$ affects our performance, then we present ablation study to evaluate the effect of each design choice. Finally, we include additional qualitative results.

\section{Implementation Details}
\label{sec:sup_implementation}

\subsection{Parameters initialization}
\label{sec:param_init}
Similar to most optimization-based methods, parameter initialization is an important factor that affects our performance. To provide good initial poses for the Gaussians, we first estimate 3D medial axes from the observed point clouds with Laplacian-Based Contraction~\cite{cao2010point,meyer2023cherrypicker}. Then we sample $m$ farthest points on the estimated medial axes as the Gaussian centers using~\cite{qi2017pointnet++}. The rotations $\mymat{R}$ are initialized as identity matrices. As for initializing the scales $s$, since scales are shared across all time steps, we find it helpful to perform a short warm up optimization stage by fitting the first set of Gaussians to the first observed point cloud with only $\mathcal{L}_{MLE} $ and $\mathcal{L}_{sep}$. 

We only apply these two parameter initialization techniques on the RoboArt dataset. The Sapien, Partial-RoboArt and Occluded-RoboArt dataset contain sparse point clouds, or point clouds with missing points. Therefore, medial axis prediction does not work well on these datsaets. For these datasets, we initialize the Gaussian centers by performing farthest point sampling directly on the input point clouds. 

\subsection{Optimization details}
We conduct our experiments on a single A100 GPU. All optimizations are done with PyTorch~\cite{Ansel_PyTorch_2_Faster_2024} and Adam optimizer~\cite{kingma2015adam}. Learning rate is set to $2e-3$ for Gaussian parameters and $1.5e-2$ for kinematic models. 
All lambdas $\lambda$ of the optimization losses are set to $1$, except that we set $\lambda_{EMD} = 0.3$. The $\alpha$ in $\mathcal{L}_{sep}$ is set to $0.5$. In every optimization iteration, we find it helpful to first perform one gradient update step with only $\mathcal{L}_{MLE}$ and $\mathcal{L}_{sep}$, followed by another gradient update step with all the losses excluding $\mathcal{L}_{MLE}$.
In the RoboArt dataset, we optimize the Gaussian parameters for $15,000$ iterations. To speed up the process, we omit $\mathcal{L}_{EMD}$ in the first $5,000$ iterations. After $5,000$ iterations, we replace $\mathcal{L}_{CD}$ with $\mathcal{L}_{EMD}$. As for the Sapien dataset, we optimize for $2,500$ iterations, while we apply $\mathcal{L}_{EMD}$ only after $1,000$ iterations. After we project the estimated part poses onto the kinematic model, we optimize the joint parameters through forward kinematics for another $500$ iterations with $\mathcal{L}_{CD}$ and $\mathcal{L}_{EMD}$.

\subsection{Losses for Partial-RoboArt and Occluded-RoboArt}

The $\mathcal{L}_{CD}$ and $\mathcal{L}_{EMD}$ defined in the main paper enforce point cloud similarity between the point clouds transformed with the estimated part segmentation and motion to match with the observed point cloud sequence. However, this supervision will not work on point cloud sequences with missing points. Since each point cloud in the sequence can have different missing parts, the point clouds will not match with each other after transformed to the same time step, even with the ground truth parameters. Therefore, we modify $\mathcal{L}_{CD}$ and $\mathcal{L}_{EMD}$ for the Partial-RoboArt and Occluded-RoboArt dataset.

In every iteration, given an input point cloud $\mypcd{X}^t$ at time step $t$, instead of transforming this point cloud to other time steps, we transform other point clouds to this time step. Let $\tilde{\mypcd{X}}^t$ denote the point cloud created by fusing all other point clouds transformed to time step $t$:

\begin{equation}
\tilde{\mypcd{X}}^t = \bigcup_{k=1, k\neq t}^K \mathcal{H}_{k \rightarrow t}
\end{equation}
where $\mathcal{H}_{k \rightarrow t}$ denotes the point cloud $\mypcd{X}^k$ at time step $k$ transformed to time step $t$ with the estimated part segmentation and motion. Due to missing points, $\tilde{\mypcd{X}}^t$ should approximate the superset of $\mypcd{X}^t$, therefore we calculate the one-directional Chamfer distance $\mathcal{L}_{CD}$ from the observed point cloud $\mypcd{X}^t$ to the fused point cloud $\tilde{\mypcd{X}}^t$:

\begin{equation}
 \mathcal{L}_{CD} =  \sum_{x \in \mypcd{X}^t} \min_{y \in \tilde{\mypcd{X}}^t} \|x - y\|_2^2
\end{equation}

Similarly, we calculate $\mathcal{L}_{EMD}$ as:

\begin{equation}
\mathcal{L}_{EMD} = \min_{A, B} \|A\mypcd{X}^t - B\tilde{\mypcd{X}}^t\|_2^2 
\end{equation}
where $A$ and $B$ are the assignment indices solved using the linear assignment solver~\cite{crouse2016implementing}. 

In addition to $\mathcal{L}_{CD}$ and $\mathcal{L}_{EMD}$ we add another loss to regularize the transformed point clouds to match with each other. Specifically, we transform all the input point clouds to the sampled time step $t$, yielding a set of point clouds: $\{\mathcal{H}_{k \rightarrow t} \, | \, k = [1,...,K]\}$. Each point cloud in this set should represent the same object pose, however, each of them may have different missing parts. Therefore, we randomly split this set into two equal sets, and create two point clouds by fusing the point clouds in each of the smaller set. Then we calculate the Chamfer distance between these two fused point clouds to serve as a regularization term.  

\subsection{Kinematic model estimation}
In this section, we include more details of how we adapt the kinematic model estimation method from~\cite{liu2023building} to our method.

\subsubsection{Part merging:}
After the Gaussian parameters are optimized in step 1, we aim to reduce over-segmentation by merging neighboring parts whose relative poses are static throughout the whole sequence. For a pair of neighboring parts $i$ and $j$, the deviation of their relative poses from a static motion is measured as:
\begin{equation}
\mathcal{L}_{merge} = \sum_k \| ({\mymat{O}_i^k}^{-1}  \cdot \mymat{O}_j^k)- \mymat{I} \|_{F}^2
\end{equation}
where $\mymat{I}$ denotes the identity matrix and $\| \cdot \|_F$ denotes the Frobenius norm. $\mymat{O}^k_i = {\mymat{T}_i^{k+1}} \cdot {\mymat{T}_i^{k}}^{-1} $ represents the pose of part $i$ in the world frame between consecutive time steps $k$ and $k+1$. Hence, $({\mymat{O}_i^k}^{-1}  \cdot \mymat{O}_j^k)$ calculates the relative motion of the two parts in the world frame at time step $k$.
We iteratively merge neighboring part pairs with $\mathcal{L}_{merge} < 3e-2$, starting from the pair with the lowest $\mathcal{L}_{merge}$, until all the remaining pairs exceed this threshold. 

\subsubsection{Kinematic tree estimation:}
We extract the kinematic tree after merging the parts that are relatively static to each other. As mentioned in the main paper, we compute pairwise $\mathcal{L}_{spatial}$ and $\mathcal{L}_{1-DOF}$ between all remaining parts. Then we estimate a minimum spanning tree such that the total loss: $\lambda_{spatial}\mathcal{L}_{spatial} + \lambda_{1-DOF}\mathcal{L}_{1-DOF}$ of all edges in the tree is minimized. In the experiments, we set $\lambda_{spatial} = 100$ and $\lambda_{1-DOF} = 1$.

$\mathcal{L}_{1-DOF}$ measures the deviation of the observed relative motion between part $(i,j)$ from the approximated ideal 1-DOF motion. $\mathcal{L}_{1-DOF}$ is calculated as:

\begin{equation}
\mathcal{L}_{1-DOF} = \sum_k \bigg\| ({\mymat{O}_i^k}^{-1}  \cdot \mymat{O}_j^k) \cdot {\mymat{S}_{ij}^k}^{-1} - \mymat{I} \bigg\|_{F}^2
\end{equation}
where ${\mymat{S}_{ij}^k}$ is the approximated 1-DOF relative motion at time step $k$. To calculate ${\mymat{S}_{ij}}$ by considering the relative poses across all time steps, we opt for the screw parameters representation used in \cite{liu2023building} to solve:

\begin{equation}
{\mymat{S}_{ij}} = \argmin_{\mymat{S}_{ij}} \bigg(  \sum_k \bigg\| ({\mymat{O}_i^k}^{-1}  \cdot \mymat{O}_j^k) \cdot {\mymat{S}_{ij}^k}^{-1} - \mymat{I} \bigg\|_{F}^2  \bigg)
\end{equation}
where ${\mymat{S}_{ij}}$ is a sequence of poses parameterized with screw parameters. Specifically, the motion is defined by a static joint location and axis (either rotation or translation axis), as well as time-dependent joint states (rotation or translation along the axis at each time step). 

After we estimate the kinematic tree structure, the joint states are initialized by approximating the screw parameters from the Gaussian poses. We directly use the screw parameter approximation and kinematic projection implemented by \cite{liu2023building} in the experiments. Finally, we fine-tune the approximated kinematic parameters with forward kinematics to minimize $\mathcal{L}_{CD}$ and $\mathcal{L}_{EMD}$.


\section{Additional Experiments and Results}
\label{sec:sup_result}

\subsection{Number of 3D Gaussians during initialization}

\begin{figure}[tb]
  \centering
  \includegraphics[width=0.65\linewidth,trim={0.2cm 0.0cm 0.5cm 2cm},clip]{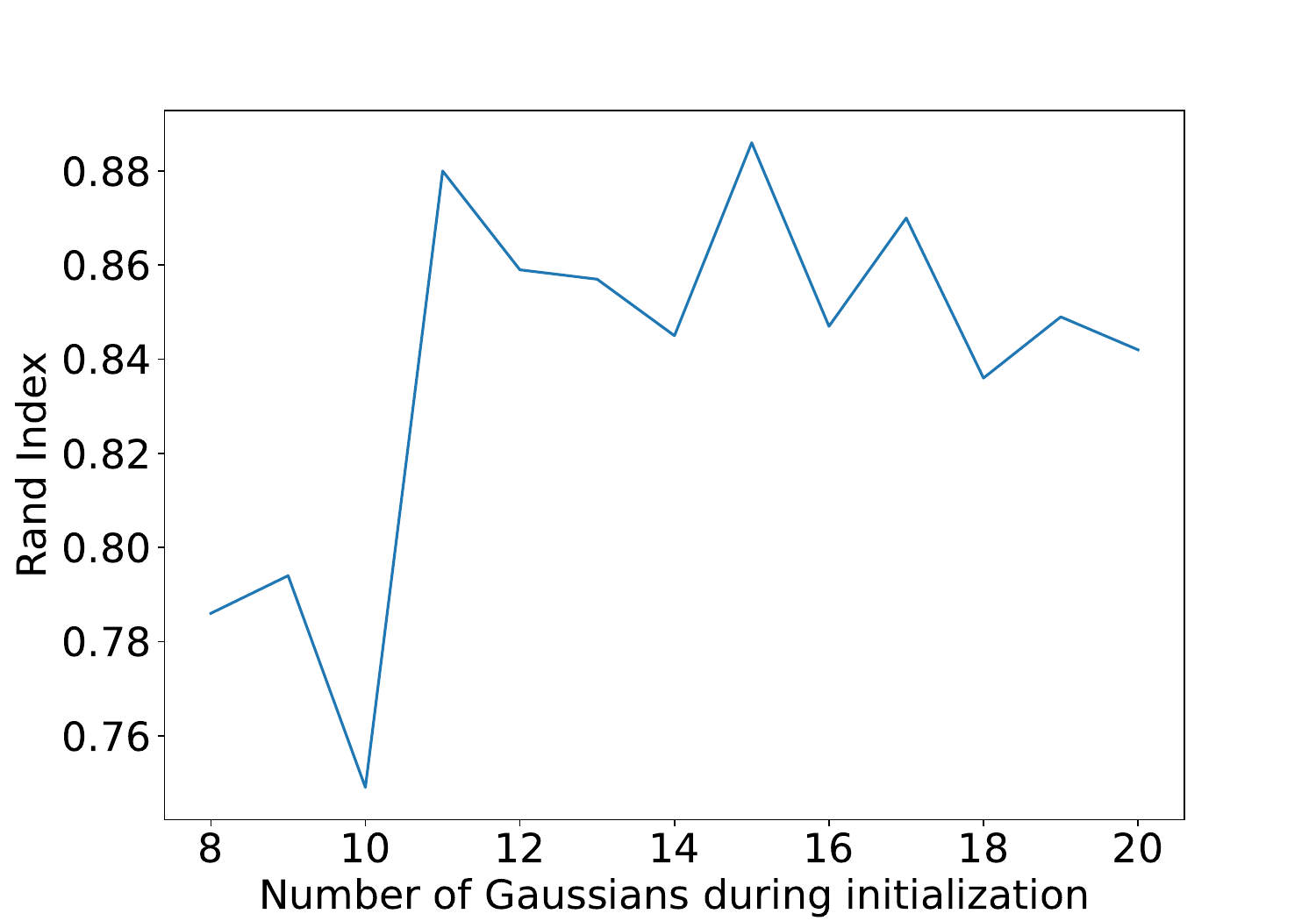}
  \caption{Part segmentation accuracy over number of initialized Gaussians on a quadruped robot with 13 rigid parts. \label{fig:plot}}
\end{figure}

To evaluate how the number of initialized Gaussians $m$ affects our performance, we ran our method with varying initializations on a quadruped robot with 13 rigid parts. As shown in Fig.~\ref{fig:plot}, the segmentation performance is less affected when the number of Gaussians exceeds the actual number of parts. This is due to the presented merging mechanism that merges neighboring parts whose relative motion falls below a predefined threshold.   In contrast, when too few Gaussians are initialized, our method can fail by being forced to represent multiple rigid parts with single a Gaussian.

In the main paper, we empirically initialize this number with several values and provide a systematic way to select the best one by choosing the initialization with the lowest reconstruction loss after optimization.
Alternatively, an off-the-shelf method~\cite{thai20243} can be used to estimate the number of parts roughly, followed by selecting a larger number of Gaussians, as our merging mechanism can handle over-initialization. We plan to investigate in leveraging a pre-trained network for part initialization in the future.

\subsection{Ablation study}


In addition to the ablation study presented in the main paper, we include the ablation study (\tabref{table:ablation}) on the two parameter initialization techniques mentioned in \secref{sec:param_init}, showing that they help improve the overall performance by providing a better initialization for optimization.

\begin{table}[t]
  \centering
  \begin{minipage}[b]{0.5\textwidth}
    \centering
    \resizebox{\columnwidth}{!}{\begin{tabular}{ccccc}
    \toprule
    Design choice       & \begin{tabular}[c]{@{}c@{}}Recons\\ Error$\downarrow$\end{tabular} & \begin{tabular}[c]{@{}c@{}}Rand\\ Index$\uparrow$\end{tabular} & \begin{tabular}[c]{@{}c@{}}Flow\\ Error$\downarrow$\end{tabular}  & \begin{tabular}[c]{@{}c@{}}Tree Edit\\ Distance$\downarrow$\end{tabular}\\ 
    \midrule
    w/o medial axes init &  2.91 & 0.87& 1.80& 5.0\\
    w/o warm up & 1.99 & 0.87 & 1.55 & 4.0 \\
    w/o $\mathcal{L}_{MLE}$  & 3.03 & 0.81 & 1.90 & 6.25   \\ 
    w/o $\mathcal{L}_{sep}$  &  1.45 & 0.88 & 1.21 & 4.0 \\
    w/o $\mathcal{L}_{CD}$   &  1.51 & 0.87 & 1.14 & 4.0 \\
    w/o $\mathcal{L}_{EMD}$  &  10.04 & 0.75 & 6.70 & 6.5  \\
    w/o $\mathcal{L}_{flow}$  &  14.52 & 0.86 & 16.19 & 4.5 \\
    Ours          & \textbf{1.26} & \textbf{0.89}   & \textbf{1.05}   & \textbf{2.75}                                  
    \end{tabular}}
    \caption{Ablation study on RoboArt validation set. We evaluate the importance of each loss term and each initialization technique by removing them one at a time.}
    \label{table:ablation}
  \end{minipage}
  \hfill
  \begin{minipage}[b]{0.48\textwidth}
    \centering
    \resizebox{\columnwidth}{!}{\begin{tabular}{lcccc}
    \toprule
     & Reart~\cite{liu2023building} & Ours${}^\dagger$ & Ours \\ 
    \midrule
    Recons Error $\downarrow$     & 7.66     & 2.68 &  \textbf{1.73}   \\
    Cano Frame RI $\uparrow$   & 0.75     & 0.85 & \textbf{0.88}      \\
    Mean RI $\uparrow$  & -     & 0.83 & \textbf{0.86}    \\
    Flow Error $\downarrow$  & 4.73     & 1.74 & \textbf{1.30}    \\
    Tree Edit Distance $\downarrow$ & 6.3     & 5.4 & \textbf{5.2} \\
    \bottomrule
    \end{tabular}}
    \caption{Additional quantitative results on Occluded-RoboArt test set. ${}^\dagger$ denotes running our method with $m=20$ (without initializing with multiple $m$). }
    \label{table:sup_occludeed}
  \end{minipage}
  
\end{table}


\subsection{Quantitative results on Occluded-RoboArt dataset.}

In the main paper, we compare our method against the baseline with the same initial $m$ on the Occluded-RoboArt dataset. 
To showcase the upper bound performance of our method on this challenging dataset, we present our result by running multiple initial $m$ and select the best $m$ after optimization in \tabref{table:sup_occludeed}.


\subsection{Qualitative results on RoboArt and Occluded-RoboArt.}
\begin{figure}[]
    \centering
    \includegraphics[width=1\linewidth,trim={0.0cm 0cm 0cm 0cm},clip]{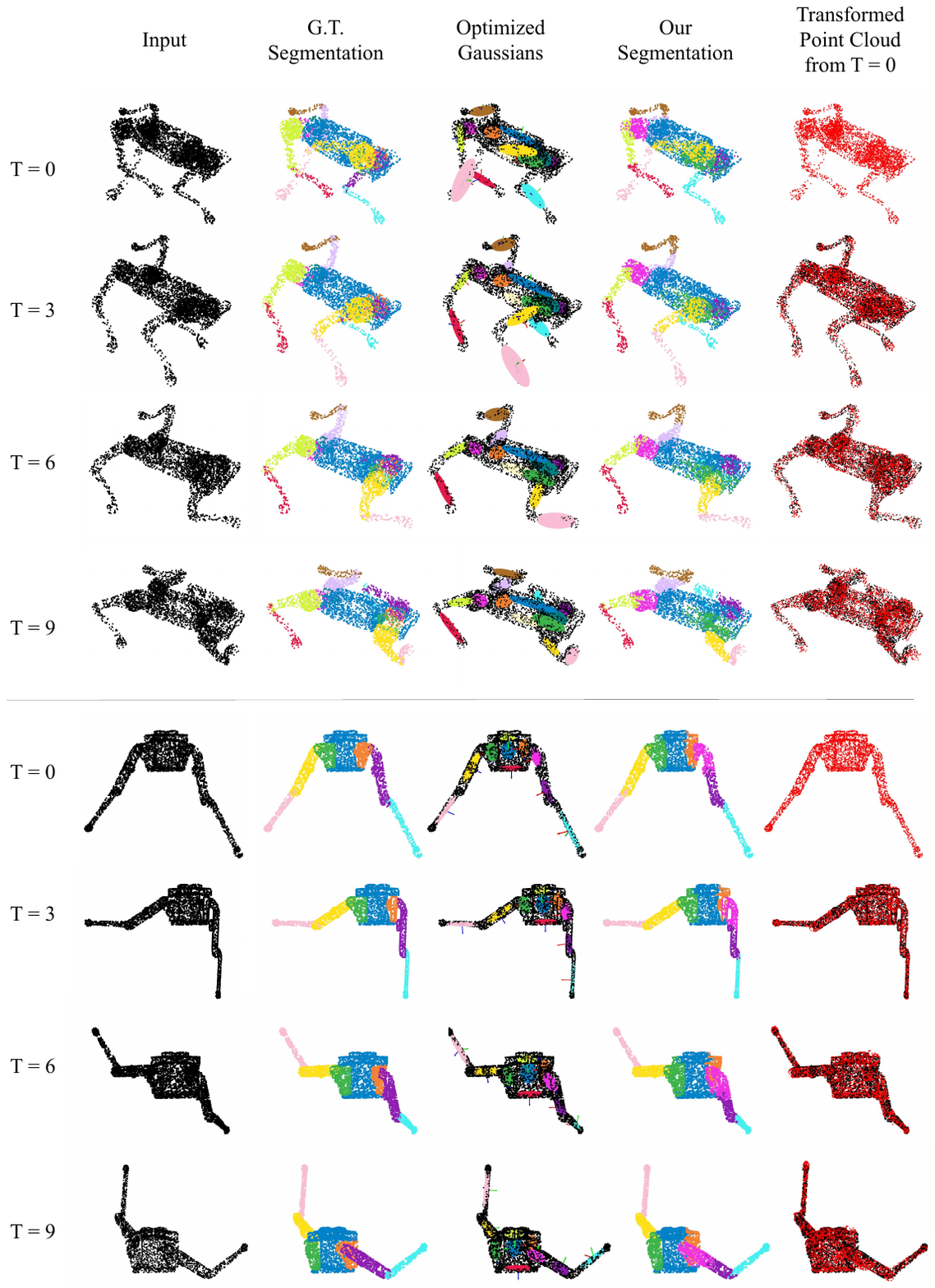}
    \caption{Additional qualitative results on RoboArt.}
    \label{fig:sub_quali_robotart_1}
\end{figure}

\begin{figure}[]
    \centering
    \includegraphics[width=1\linewidth,trim={0.0cm 0cm 0cm 0cm},clip]{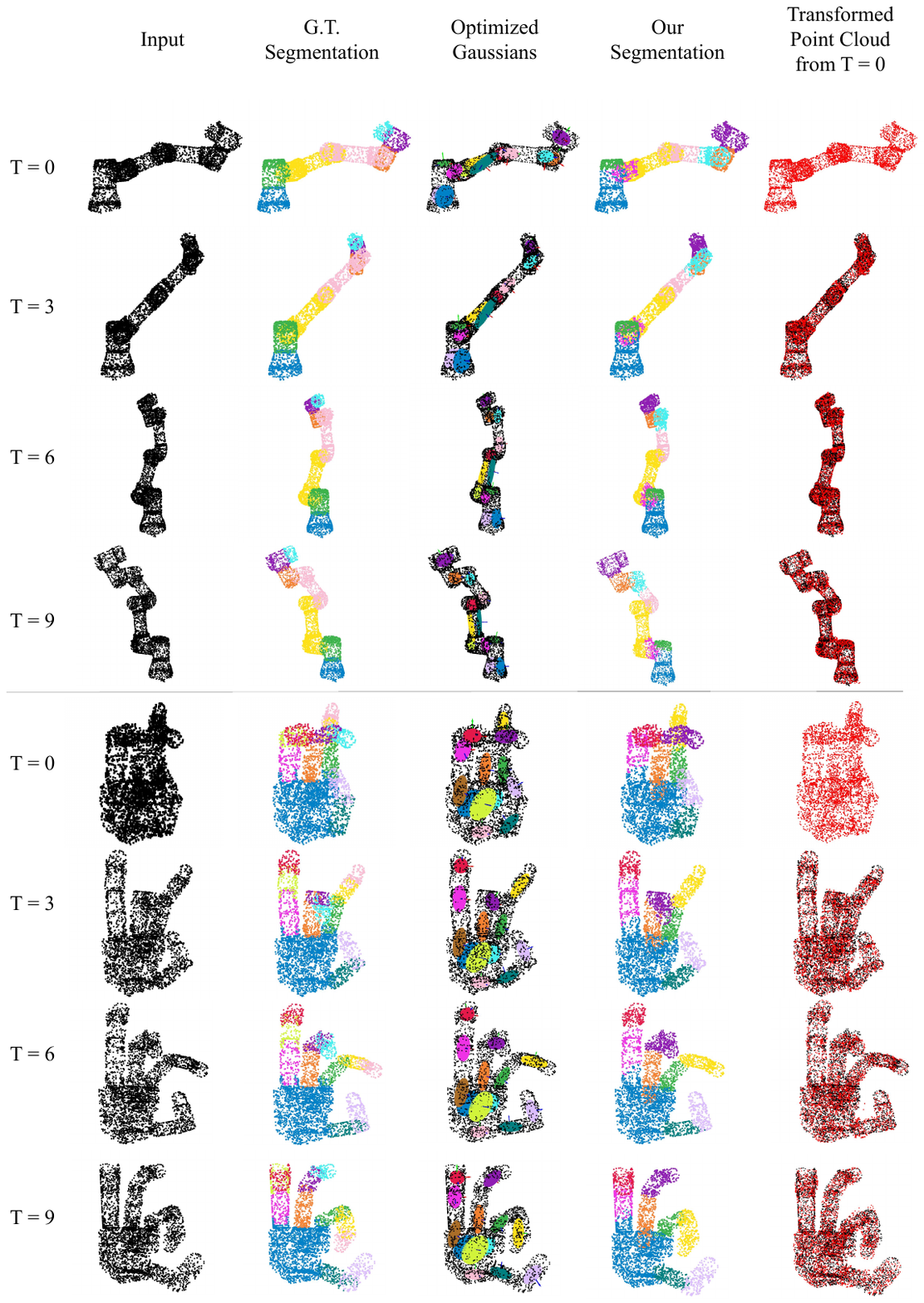}
    \caption{Additional qualitative results on RoboArt.}
    \label{fig:sub_quali_robotart_2}
\end{figure}

\begin{figure}[]
    \centering
    \includegraphics[width=1\linewidth,trim={0.0cm 0cm 0cm 0cm},clip]{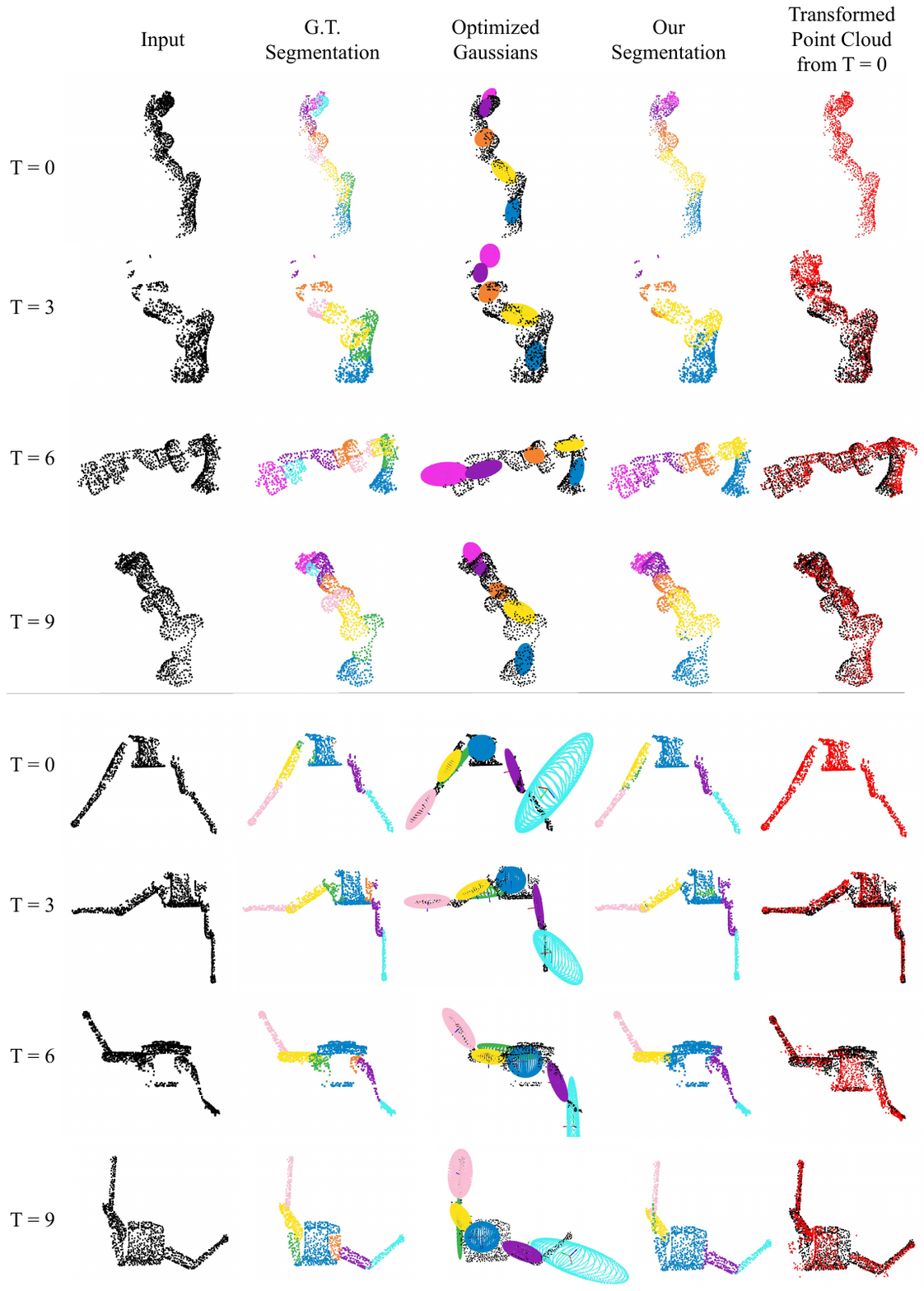}
    \caption{Additional qualitative results on Occluded-RoboArt. Note that each point cloud in the sequence contains different missing parts, therefore the transformed points from $T=0$ (red) will not perfectly overlap with the observed points at each time step (black) even with the ground truth parameters. However, we show that with our estimated motion, the union of the red and black points cover the surface of the object faithfully. }
    \label{fig:sub_quali_occluded_1}
\end{figure}

Finally, we present additional qualitative results on the RoboArt dataset in \figref{fig:sub_quali_robotart_1}, \figref{fig:sub_quali_robotart_2} and Occluded-RoboArt in \figref{fig:sub_quali_occluded_1}. 
In these figures, the observed point clouds and the ground truth part segmentation are presented in column 1 and 2, respectively. In column 3, we visualize the 3D Gaussians after optimization, while column 4 shows the predicted part segmentation. 
Moreover, to demonstrate the result of the predicted part motions, we overlap the points transformed from $T=0$ to the other time steps (red) and the observed point clouds at the respective time steps (black). 
Note that each point cloud is arbitrarily sampled from the object surface at each time step. Therefore, the red points and the black points will not align perfectly even with the ground truth parameters. Nevertheless, we demonstrate that the union of the red and black points covers the object surface faithfully. More visualization of the results can be found in the video. 

\newpage
\bibliography{references}